\documentclass[10pt,twocolumn,letterpaper]{article}%

\usepackage{cvpr}
\usepackage{times}
\usepackage{epsfig}
\usepackage{graphicx}
\usepackage{amsmath}
\usepackage{amssymb}

\usepackage[labelfont=bf]{caption}
\usepackage[table]{xcolor}
\definecolor{Gray}{gray}{0.85}
\usepackage{color}

\usepackage{booktabs}
\usepackage{multirow}

\usepackage[pagebackref=true,breaklinks=true,letterpaper=true,colorlinks,bookmarks=false]{hyperref}

\cvprfinalcopy 

\def\cvprPaperID{2244} 

\begin{document}

\title{Effective Face Frontalization in Unconstrained Images}

\author{
\begin{tabular}{c@{\extracolsep{0.6cm}}c@{\extracolsep{0.6cm}}c@{\extracolsep{0.6cm}}c}
Tal Hassner$^{1}$ & Shai Harel$^{1~\dagger}$ & Eran Paz$^{1~\dagger}$ & Roee Enbar$^{2}$ \\
\end{tabular}\\
{\small $^{1}$ The open University of Israel}\\
{\small $^{2}$ Adience}
\vspace{-0.5cm}
}

\makeatletter
\let\@oldmaketitle\@maketitle
\renewcommand{\@maketitle}{\@oldmaketitle
  \vspace{-0.5cm}\includegraphics[width=\linewidth,clip,trim = 0mm 0mm 0mm 0mm]{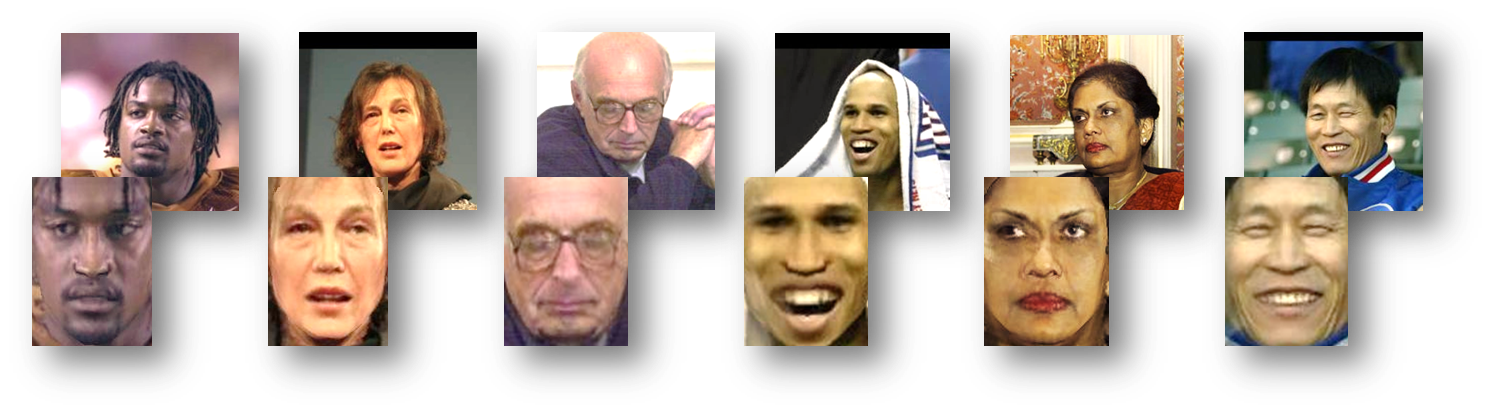}\captionof{figure}{{\bf Frontalized faces.} Top: Input photos; bottom: our frontalizations, obtained without estimating 3D facial shapes.}\bigskip}
\makeatother

\maketitle

\begin{abstract}
``Frontalization'' is the process of synthesizing frontal facing views of faces appearing in single unconstrained photos. Recent reports have suggested that this process may substantially boost the performance of face recognition systems. This, by transforming the challenging problem of recognizing faces viewed from unconstrained viewpoints to the easier problem of recognizing faces in constrained, forward facing poses. Previous frontalization methods did this by attempting to approximate 3D facial shapes for each query image. We observe that 3D face shape estimation from unconstrained photos may be a harder problem than frontalization and can potentially introduce facial misalignments. Instead, we explore the simpler approach of using a single, unmodified, 3D surface as an approximation to the shape of all input faces. We show that this leads to a straightforward, efficient and easy to implement method for frontalization. More importantly, it produces aesthetic new frontal views and is surprisingly effective when used for face recognition and gender estimation.
\end{abstract}

\section{Introduction}\let\thefootnote\relax\footnote{\noindent $^\dagger$ These authors ordered alphabetically due to equal contribution.\\$^*$ For code and data, please see project page at~\url{www.openu.ac.il/home/hassner/projects/frontalize}}
Face recognition performances, reported as far back as~\cite{phillips2007frvt}, have shown computer vision capabilities to surpass those of humans. Rather than signaling the end of face recognition research, these results have led to a redefinition of the problem, shifting attention from highly regulated, controlled image settings to faces captured in unconstrained conditions (a.k.a., ``in the wild'').

This change of focus, from constrained to unconstrained images, has toppled recognition rates (see, e.g., the original results~\cite{huang2007unsupervised} on the Labeled Faces in the Wild~\cite{lfw} benchmark, published in the same year as~\cite{phillips2007frvt}). This drop was not surprising: Unconstrained photos of faces represented a myriad of new challenges, including changing expressions, occlusions, varying lighting, and non-frontal, often extreme poses. Yet in recent years recognition performance has gradually improved to the point where, once again, claims are being made for super-human face recognition capabilities (e.g.,~\cite{lu2014surpassing,sun2014deep,taigman2013deepface}).

\begin{figure*}[t!]
\includegraphics[width=\textwidth,clip,trim = 0mm 0mm 0mm 0mm]{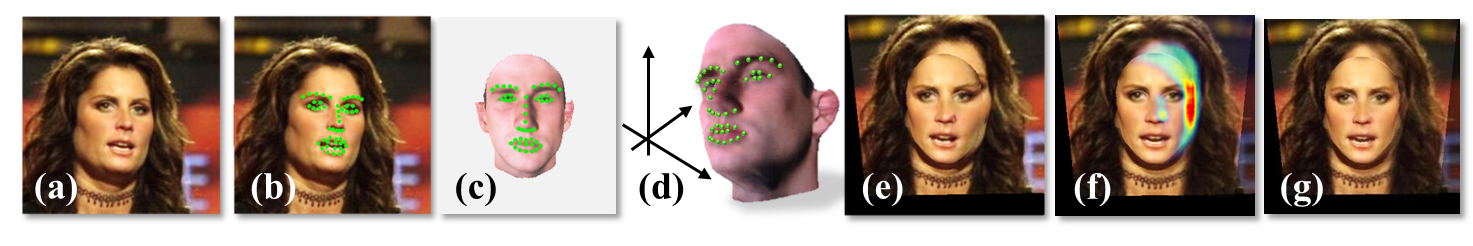}
\caption{{\bf Frontalization process overview.} (a) Query photo; (b) facial feature detections; (c) the same detector used to localize the same facial features in a reference face photo, produced by rendering a textured 3D computer graphics model (d); (e) from the 2D coordinates on the query and their corresponding 3D coordinates on the model we estimate a projection matrix which is then used to back-project query intensities to the reference coordinate system; (f) estimated visibility due to non-frontal poses, overlaid on the frontalized result. Warmer colors reflect less visible pixels. Facial appearance in these regions is produced by borrowing colors from corresponding symmetric parts of the face; (g) our final frontalized result.} \label{fig:process}
\end{figure*}

Modern methods vary in how they address the many challenges of unconstrained face recognition. Facial pose variations in particular have often been considered by designing representations that pool information over large image regions, thereby accounting for possible misalignments due to pose changes (e.g.,~\cite{lu2014surpassing,simonyan2013fisher,WHT:ECCVW08:DBMW,wolf2011effective}), by improving 2D face alignment accuracy~\cite{huang2012learning,huang2007unsupervised,WHT:ACCV09:SSBS}, or by using massive face collections to learn pose-robust representations~\cite{hu2014discriminative,sun2013hybrid,sun2014deepB,zhu2014recover}.

Recently, some proposed to simplify unconstrained face recognition by reducing it, at least in terms of pose variations, to the simpler, constrained settings. This, by automatic synthesis of new, frontal facing views, or ``frontalization''~\cite{hassner2013viewing,taigman2013deepface}. To this end, they attempt to estimate a rough approximation for the 3D surface of the face and use this surface to generate the new views. Although appealing, this approach relies on accurate localization of facial feature points and does not guarantee that the same alignment (frontalization) will be applied to different images of the same face. Thus, different images of the same person may well be aligned differently, preventing their features from being accurately compared.

We propose the simple alternative approach of using a single, unmodified 3D reference for all query faces in order to produce frontalized views. Ignoring individual differences in facial shapes may be counter-intuitive -- indeed, previous work has emphasized its importance~\cite{taigman2013deepface} -- however, qualitative examples throughout this paper show that any impact this has on facial appearances is typically negligible. In fact, faces remain easily recognizable despite this approximation. More importantly, our frontalized faces are aggressively aligned thereby improving performances over previous alignment methods. We verify our claims using the LFW benchmark for face recognition and the Adience benchmark for gender estimation. Finally, we encourage reproduction of our results and make our frontalization code and frontalized images publicly available.

\section{Related work}\label{sec:related}
Generating novel views of a face viewed in a single image has been a longstanding challenge in computer vision, due in large part to the potential applications such methods have in face processing and recognition systems.

Previous methods for synthesizing new facial views typically did so by estimating the 3D surface of the face appearing in the photo with varying emphasis on  reconstruction accuracy. Morphable-Models based methods~\cite{blanz2004exchanging,blanz1999morphable,tang2008real,yang2011expression} attempt to learn the space of allowable facial geometries using many aligned 3D face models. These methods, however, typically require near-frontal views of clear, unoccluded faces, and so not suitable for our purposes.

Shape from shading methods have been shown to produce outstanding facial details (e.g.,~\cite{kemelmacher20113d}). Their sensitivity to occlusions and specularities (e.g., eyeglasses) and requirement for careful segmentation of faces from their backgrounds make them less suited for automatic, large scale application in face processing systems. 

Facial symmetry was used in~\cite{dovgard2004statistical} to estimate 3D geometry. Like us, symmetry was used for replacing details in out-of-view facial regions in~\cite{gonzalez2007symmetry}. These methods have only been applied to controlled views due to their reliance on accurate segmentation.

Related to our work is~\cite{hassner2013viewing} and its extension for face recognition in~\cite{taigman2013deepface}. Both attempt to adjust a 3D reference face, fitting it to the appearance of the query face in order to preserve natural appearances. This 3D estimation process, however, cannot guarantee that a similar shape would be produced for difference images of the same face. It further either relies on highly accurate facial feature localizations~\cite{taigman2013deepface}, which can be difficult to ensure in practice, or is computationally heavy, unsuited for mass processing~\cite{hassner2013viewing}. We discuss these issues at length in Sec.~\ref{sec:discuss}.

Finally,~\cite{zhu2014recover} described a deep-learning based method for estimating canonical views of faces. Their method is unique in producing frontal views without estimating (or using) 3D information in the process. Besides requiring substantial training, their canonical views are not necessarily frontalized faces and are not guaranteed to be similar to the person appearing in the input image. 

We propose to use a single 3D reference surface, unchanged, in order to produce front facing views for all query images. Despite the simplicity of this approach, we are unaware of previous reports of its use in unconstrained face photo alignment for face recognition. We explore the implications of our approach both qualitatively and empirically.

\section{Hard frontalization}\label{sec:front}
We use the term ``hard frontalization'' to emphasize our use of a single, 3D, reference face geometry. This, in contrast to others who estimate or modify 3D facial geometry to fit facial appearances (Sec.~\ref{sec:related}). Our goal is to produce better aligned images which allow for accurate comparison of local facial features between different faces. As we next show, the use of a single 3D face results in a straightforward frontalization method which, despite its simplicity, is shown to be quite effective.

Our method is illustrated in Fig.~\ref{fig:process}. A face is detected using an off-the-shelf face detector~\cite{viola2004robust} and then cropped and rescaled to a standard coordinate system. The same dimensions and crop ratios previously used for Labeled Faces in the Wild (LFW)~\cite{lfw} images are used here in order to maintain parameter comparability with previous results. 

Facial feature points are localized and used to align the photo with a textured, 3D model of a generic, reference face. A rendered, frontal view of this face provides a reference coordinate system. An initial frontalized face is obtained by back-projecting the appearance (colors) of the query photo to the reference coordinate system using the 3D surface as a proxy. A final result is produced by borrowing appearances from corresponding symmetric sides of the face wherever facial features are poorly visible due to the query's pose. These steps are details next.

\subsection{Generating a frontalized view}\label{sec:pose}
We begin by computing a $3\times 4$ projection matrix which approximates the one used to capture the query photo. To this end, we seek 2D-3D correspondences between points in the query photo (Fig~\ref{fig:process} (a)) and points on the surface of our 3D face model (Fig~\ref{fig:process} (d)). This, by matching query points to points on a rendered, frontal view of the model (Fig~\ref{fig:process} (c)). Directly estimating correspondences between a real photo and a synthetic, rendered image can be exceedingly hard~\cite{hassner2014standard}. Instead, we use a robust facial feature detection method which seeks the same landmarks (e.g., corners of the eyes, mouth etc.) in both images.\\

\noindent{\bf Facial feature detection.} Many highly effective methods were recently proposed for detecting facial features. In designing our system, we tested several state-of-the-art detectors, selecting the supervised descent method (SDM) of~\cite{xiong2013supervised} as the one which balances both speed of detection with accuracy. Unlike others, the 48 facial features it localizes do not include points along the jawline (Fig~\ref{fig:process} (b-c)). The features it detects are therefore all images of points lying close to the 3D plane at the front of the face. These and other concerns have been suggested in the past as reasons for preferring other approaches to pose estimation~\cite{cao2014displaced,saragih2011real}. As we later show, this did not appear to be a problem in our tests. \\

\noindent{\bf Pose estimation.} Given a textured 3D model of a face, the synthetic, rendered view of this model is produced by specifying a reference projection matrix $\mathbf{C}_M = \mathbf{A}_M~~[\mathbf{R}_M~~\mathbf{t}_M]$, where $\mathbf{A}_M$ is the intrinsic matrix, and $[\mathbf{R}_M~~\mathbf{t}_M]$ the extrinsic matrix consisting of rotation matrix $\mathbf{R}_M$ and translation vector~$\mathbf{t}_M$. We select rotation and translation to produce a frontal view of the model (Fig.~\ref{fig:process} (c)) which serves as our reference (frontalized) coordinate system.

When producing the reference view $I_R$ we store for each of its pixels $\mathbf{p}'$ the 3D point coordinates $\mathbf{P}=(X,Y,Z)^T$ of the point located on the surface of the 3D model for which:
\begin{equation}
\mathbf{p}'\sim \mathbf{C}_M ~~\mathbf{P}.\label{eq:proj_ref}
\end{equation}
\noindent Let $\mathbf{p}_i=(x_i,y_i)^T$ be facial feature points detected in the query photo $I_Q$ (Fig.~\ref{fig:process} (b)), and $\mathbf{p}'_{i}=(x'_i, y'_i)^T$ be the same facial features, detected in the reference view (Fig.~\ref{fig:process} (c)). From Eq.~\ref{eq:proj_ref}, we have the coordinates $\mathbf{P}_i = (X_i, Y_i, Z_i)^T$ of the point on the surface of the model, projected onto $\mathbf{p}'_{i}$ (Fig.~\ref{fig:process} (d)). This provides the correspondences $(\mathbf{p}_i^T,\mathbf{P}_i^T) = (x_i, y_i, X_i, Y_i, Z_i)$ which allow estimating the projection matrix $\mathbf{M}_Q = \mathbf{A}_Q~~[\mathbf{R}_Q~~\mathbf{t}_Q]$, approximating the camera matrix used to capture the query photo $I_Q$~\cite{Hartley2004}. Projection matrix estimation itself is performed using standard techniques (Sec.~\ref{sec:results}).\\

\noindent{\bf Frontal pose synthesis.} An initial frontalized view $I_F$ is produced by projecting query facial features back onto the reference coordinate system using the geometry of the 3D model. For every pixel coordinate $\mathbf{q}'=(x',y')^T$ in the reference view, from Eq.~\ref{eq:proj_ref} we have the 3D location $\mathbf{P}=(X,Y,Z)^T$ on the surface of the reference which was projected onto $\mathbf{q}'$ by $\mathbf{C}_M$. We use the expression 
\begin{equation}
\mathbf{p}\sim \mathbf{C_Q} ~~\mathbf{P}\label{eq:proj_query}
\end{equation}
\noindent to provide an estimate for the location $\mathbf{p}=(x,y)^T$ in $I_Q$ of that same facial feature. Bi-linear interpolation is used to sample the intensities of the query photo at $\mathbf{p}$. The sampled color is then assigned to pixel coordinates $\mathbf{q}'$ in the new, frontalized view (Fig.~\ref{fig:process} (e)).

\subsection{Soft symmetry}\label{sec:symmetry}

\begin{figure}[t!]
\includegraphics[width=\linewidth,clip,trim = 0mm 0mm 0mm 0mm]{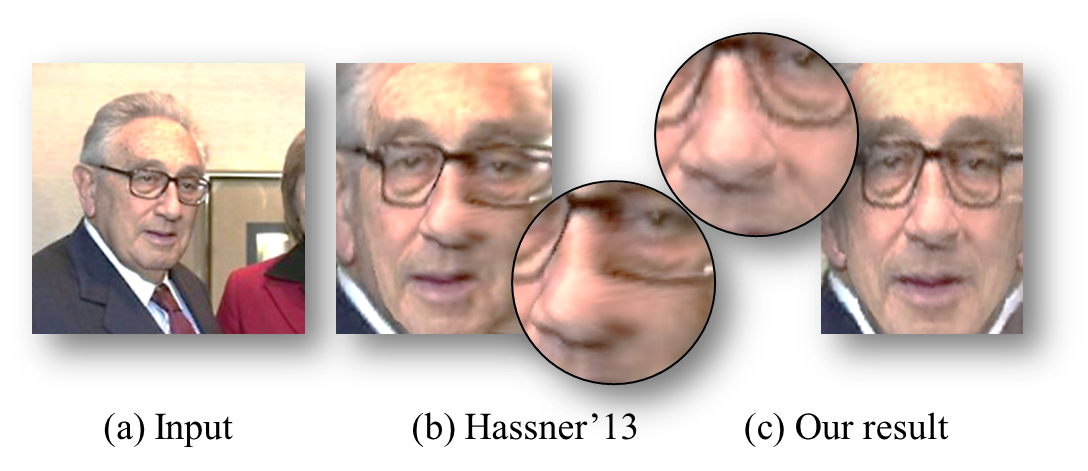}
\caption{{\bf Occlusion handling comparison.} (a) Input image. (b) Frontalization obtained by the method of~\cite{hassner2013viewing}, showing noticeable smearing artifacts wherever input facial features were occluded. (c) Our result, produced with occlusion detection and soft facial symmetry.} \label{fig:occlusions}
\end{figure}

Out-of-plane rotation of the head can cause some facial features to be less visible than others, particularly those on the sides of the nose and head. In~\cite{hassner2013viewing} a depth map was used to generate new views. This has the effect of over-sampling, or ``smearing'' textures whenever they were occluded in the original photo (Fig.~\ref{fig:occlusions} (b). In~\cite{taigman2013deepface}, the authors suggest using mesh triangle visibility, presumably using 3D surface normals computed on their low resolution 3D shape estimate (Fig.~\ref{fig:surfacecompare} (top right)), though it is not clear if they employed this approach in practice. In doing so they rely on the accuracy of their facial feature detector to define the exact positions of their 3D triangles. In addition, the coarse triangulation used to represent their 3D model may not provide accurate enough, per-pixel visibility estimates.\\

\noindent{\bf Estimating visibility.} We estimate visibility using an approach similar to the one used by multi-view 3D reconstruction methods (e.g.,~\cite{kutulakos2000theory,zeng2005progressive}). Rather than using two or more views to estimate 3D geometry we use an approximation to the 3D geometry (the reference face) and a single view ($I_R$) to estimate visibility in a second image ($I_Q$).

\begin{figure}[t!]
\centering{
\includegraphics[width=\linewidth,clip,trim = 0mm 0mm 0mm 0mm]{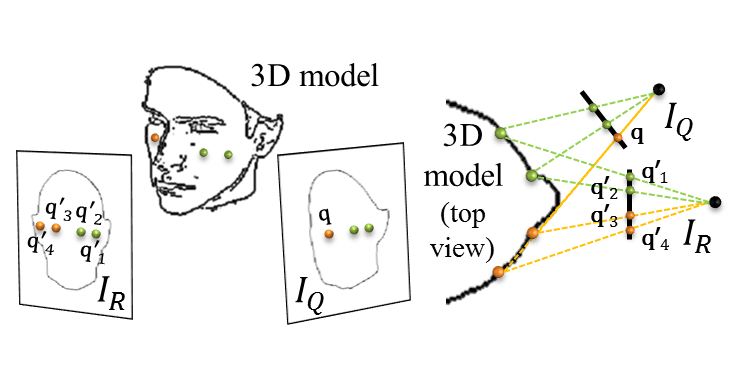}
}
\caption{{\bf Visibility estimation.} Pixels $\mathbf{q}'_3$ and $\mathbf{q}'_4$ in the reference (frontalized) coordinate system $I_R$, both map to the same pixel $\mathbf{q}$ in the query photo $I_Q$, and so would both be considered less visible. Their corresponding symmetric pixels $\mathbf{q}'_1$ and $\mathbf{q}'_2$ are used to predict their appearance in the final frontalized view.} \label{fig:explainocclusions}
\end{figure}

We evaluate visibility by counting the number of times query pixels are accessed when generating the new view: As a face rotates away from the camera, the angle between its less visible features and the camera plane increases, consequently increasing the number of surface points projected onto the same pixel in the photo (Fig.~\ref{fig:explainocclusions}). This translates to the following sampling-rate measure of visibility as follows. 

\begin{figure}[t!]
\includegraphics[width=\linewidth,clip,trim = 0mm 0mm 0mm 0mm]{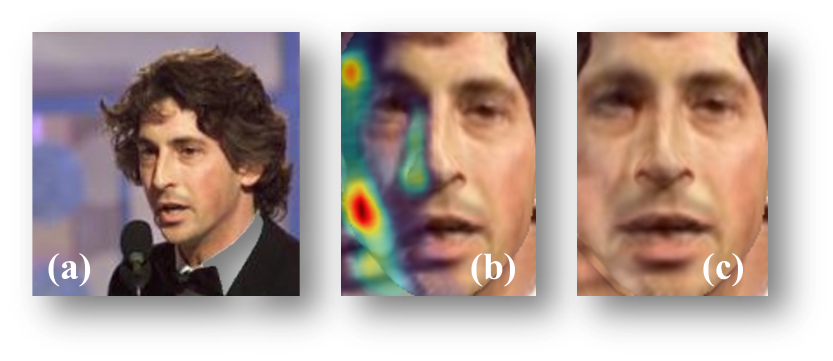}
\caption{{\bf Visibility estimation for an extreme out-of-plane pose.} (a) Input image. (b) Visibility estimates overlaid on the initial frontalized image. Warmer colors reflect less visibility of these features in the original input image (a). (c) Frontalization with soft-symmetry.} \label{fig:occlusionmap}
\end{figure}

Returning to Eq.~\ref{eq:proj_query}, for each pixel $\mathbf{q}'$ in the reference view $I_R$, we store the location in the query photo of its corresponding pixel $\mathbf{q}$ (in practice, a quantized, integer value reflecting the nearest neighboring pixel for the non-integer value of $\mathbf{q}$). A visibility score is then determined for each pixel in the frontalized $\mathbf{q'}$ view by:
\begin{equation}
v(\mathbf{q'}) = 1- \exp(- \#\mathbf{q}).\label{eq:visibility}
\end{equation}
\noindent Where $\#\mathbf{q}$ is the number of times query pixel $\mathbf{q}$ corresponded with {\em any} frontalized pixel $\mathbf{p}'$. Fig.~\ref{fig:process} (f) and Fig.~\ref{fig:occlusionmap} (b) both visualize the estimated visibility rates for two faces, overlaid on the initial frontalized results. In both cases, facial features turned away from the camera are correctly highlighted. 

We note that an alternative method of projecting the surface normals of the 3D model down to the query photo and using their values to determine visibility can also be employed. We found the approach described above faster and both methods provided similar results in practice.

Intensities of poorly visible pixels (low visibility scores in Eq.~\ref{eq:visibility}) are replaced by a mean of their intensities and the intensities of their corresponding symmetric pixels, weighted by the visibility scores. We note that this weighing of symmetric parts of the face can produce artifacts, especially when the head is at non-frontal poses and lighting on both parts of the face are different (e.g., the example in Fig.~\ref{fig:occlusionmap}). Although more elaborate methods of blending the two parts of the face can be employed, descriptors commonly used for face recognition are typically designed to overcome noise and minor artifacts such as these, and so other blending methods were not used here.

\subsection{Conditional soft-symmetry} \label{sec:conditional}
\begin{figure}[t!]
\includegraphics[width=\linewidth,clip,trim = 0mm 0mm 0mm 0mm]{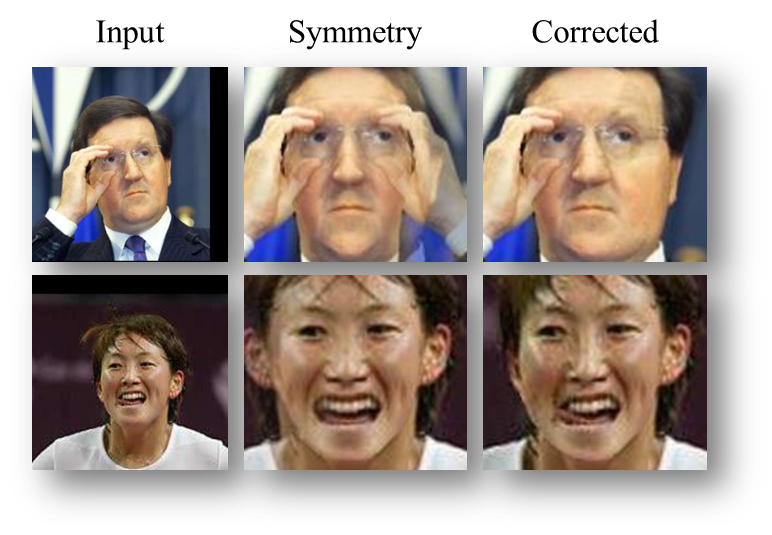}
\caption{{\bf Corrected soft-symmetry examples.} Left: Input image; Mid: results following soft symmetry. Right: Non-symmetric results, automatically selected due to detected symmetry errors. In the top row, symmetry replicated an occluding hand, in the bottom an unnatural expression was produced by transferring an asymmetric expression.} \label{fig:fixocclude}
\end{figure}

Although transferring appearances from one side of the face to another may correct pose related visibility issues, it can also introduce problems whenever one side of the face is occluded by anything other than the face itself: symmetry can replicate the occlusion, leaving the final result unrecognizable. Asymmetric facial expressions, lighting, and facial features may also cause frontalization errors. Two such examples are presented in Fig.~\ref{fig:fixocclude} (mid).

In order to detect these failures, we take advantage of the aggressive alignment of the frontalized images. By using the same 3D reference, features on frontalized faces appear in the same image locations regardless of the actual shape of the face. For example, the same image region in the frontalized face will always correspond to the right side of the mouth. These local appearances, following such alignment, can be easily verified using a standard robust representation and a classifier trained on example patches extracted at the same facial locations.

In our implementation, we manually specified eight location on the reference face, corresponding to the sides of the mouth, nose and eyes. We then trained eight linear SVM classifiers, one for each point, to recognize local appearances at each point, represented as LBP code~\cite{LBP2,ojala2002multiresolution}. Training examples were generated from frontalized images (in practice, LFW~\cite{lfw} images not included in the benchmark tests) using LBP code patches extracted at these eight locations from all training images.

Given a new frontalized face, we classify its patches, extracted from the same eight locations. A frontalized face with soft symmetry is rejected in favor of the non-symmetric frontalization if more of the latter's points were correctly identified by their classifiers. Fig.~\ref{fig:fixocclude} shows two examples with (erroneous) soft symmetry and the automatically selected, non-symmetric frontalized result.

\begin{figure}[t!]
\includegraphics[width=\linewidth,clip,trim = 0mm 0mm 0mm 0mm]{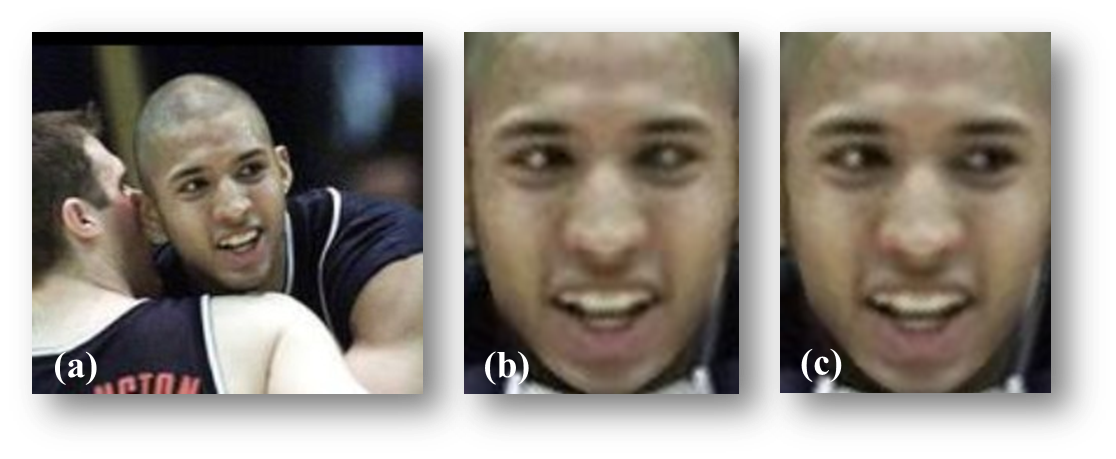}
\caption{{\bf Eye correction.} (a) Input image. (b) Frontalization with soft-symmetry. (c) Frontalization with eye-excluded soft-symmetry.} \label{fig:eyes}
\end{figure}

Finally, eyes are ignored when symmetry is applied; their appearance is unchanged from the initial frontalized view regardless of their visibility. This is done for aesthetic reasons: As demonstrated in Fig.~\ref{fig:eyes}, simply using symmetry can result in unnaturally looking, cross-eyed faces, though this exclusion of the eyes did not seem to affect our face recognition performance one way or another. To exclude the eyes from the symmetry, we again exploit the strong alignment: Eye locations are selected once, in the reference coordinate system, and the same pixel coordinates were always excluded from the soft symmetry process. \\

\section{Discussion: Soft vs. hard frontalization}\label{sec:discuss}
Unlike previous methods we do not try to tailor a 3D surface to match the appearance of each query face. Ostensibly, doing so allowed previous methods to better preserve facial appearances in the new, synthesized views. We claim that this may actually be unnecessary and possibly even counterproductive, {\em damaging}, rather than improving face recognition performance.

\begin{figure}[t!]
\centering{
\includegraphics[width=.9\linewidth,clip,trim = 5mm 3mm 5mm 5mm]{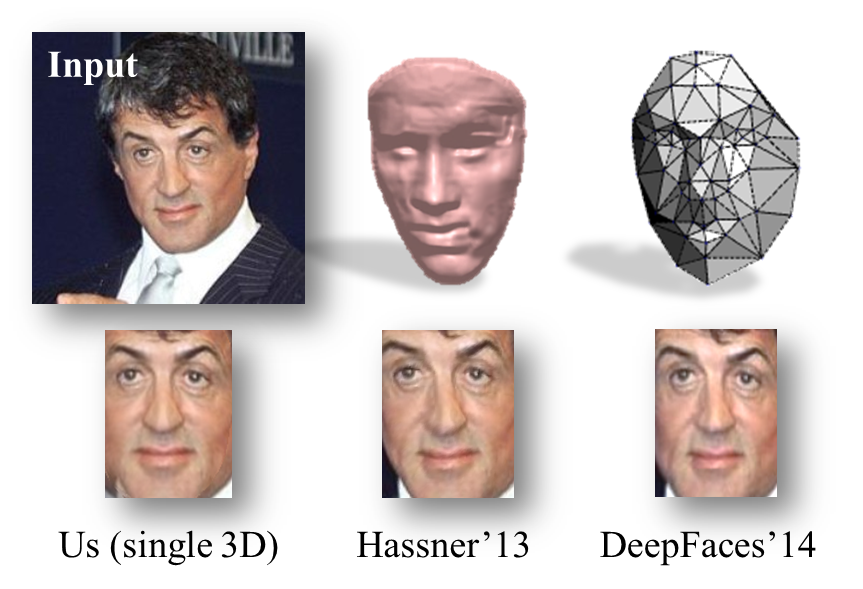}
}
\caption{{\bf Visualization of estimated 3D surfaces.} Top: Surfaces estimated for the same input image (left) by Hassner~\cite{hassner2013viewing} (mid) and DeepFaces~\cite{taigman2013deepface} (right). Bottom: Frontalized faces using our single-3D approach (left), Hassner (mid) and DeepFaces (right). Evidently, both surfaces are very rough approximations to the shape of the face. Moreover, despite the different surfaces, all three results seem qualitatively similar. This calls to question the need for shape estimation or fitting when performing frontalization.\vspace{-3mm}} \label{fig:surfacecompare}
\end{figure}

In~\cite{taigman2013deepface}, 3D facial geometry was altered by using the coordinates of detected facial feature points to modify a 3D surface, matching it to the query face. This surface, however, is a rough approximation of the true facial geometry which preserves little if any identifying features (Fig.~\ref{fig:surfacecompare} (top-right)). Furthermore, there is no guarantee that local feature detections will be repeatedly detected in the same exact positions in different views of the same face. Thus, different shapes could be estimated for different views of the same face, resulting in misaligned features and possible noise.

Although the problem of accurately detecting facial feature points is somewhat ameliorated in~\cite{hassner2013viewing} by using dense correspondences rather than sparse image detections, they too produce only a rough approximation of the subject's face (Fig.~\ref{fig:surfacecompare} (top-mid)) and similarly cannot guarantee alignment of the same facial features across different images.

Of course, face shape differences may provide important cues for recognition. This is supported by many previous reports~\cite{farkas1981anthropometry} which have found significant age, gender and ethnicity based differences in facial shapes. However, previous frontalization methods do not guarantee these differences will actually be preserved, implicitly relying on texture rather than shape for recognition. This is evident in Fig.~\ref{fig:surfacecompare} (bottom), where frontalizations for these two methods and our own appear qualitatively comparable.

\begin{figure}[t!]
\includegraphics[width=\linewidth,clip,trim = 0mm 0mm 0mm 0mm]{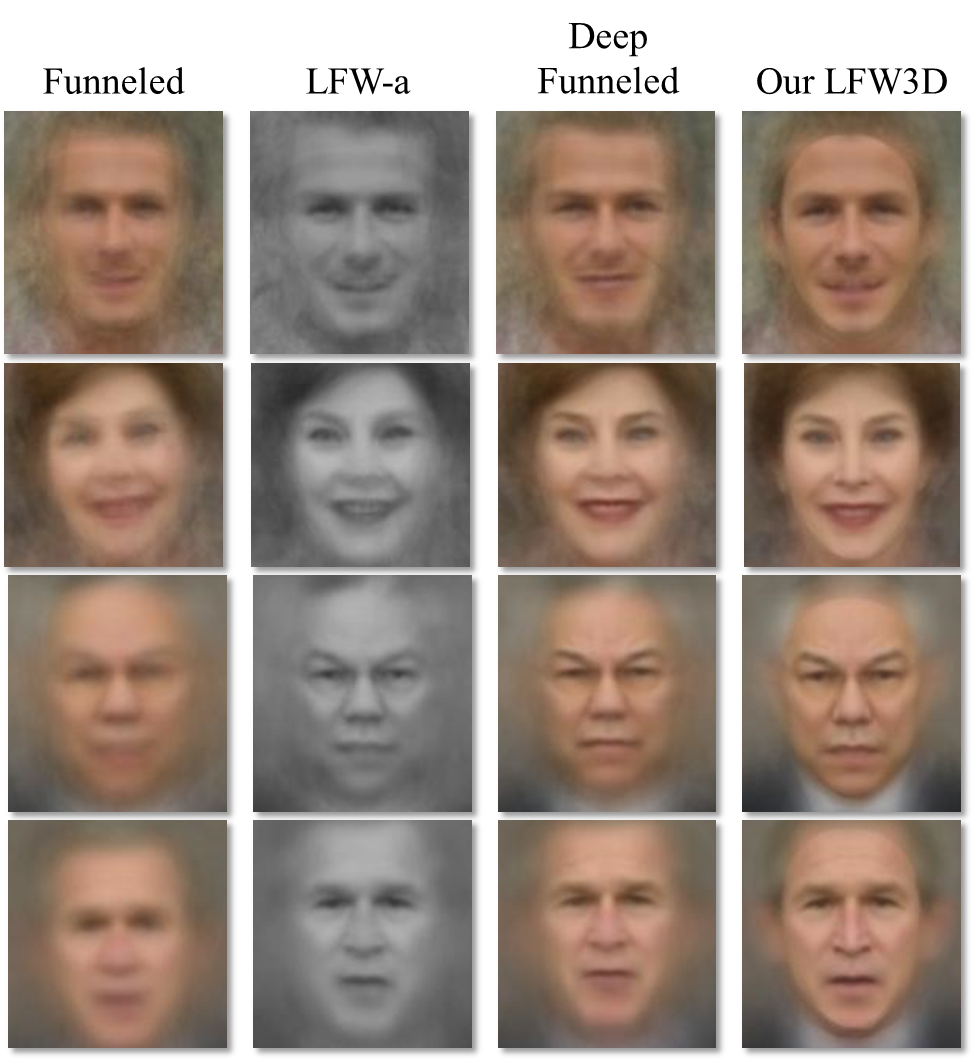}
\caption{{\bf Mean faces with different alignment methods.} Average faces from the 31 David Beckham, 41 Laura Bush, 236 Colin Powell, and 530  George W. Bush images in the LFW set. From left to right, columns represent different alignments: The Funneling of~\cite{huang2007unsupervised}, the LFW-a images (available only in grayscale)~\cite{WHT:ACCV09:SSBS}, the deep-funneled images of~\cite{huang2012learning} and our own frontalized faces. Wrinkles on the forehead of George W Bush in our result are faintly visible. These were preserved despite having been averaged from 530 images captured under extremely varying conditions.} \label{fig:meanfaces}
\end{figure}

Observing that for frontalization, one rough approximation to the 3D facial shape seems as good as another, we propose using the same 3D reference, unmodified with all faces. In doing so, we abandon attempts to preserve individual 3D facial structures in favor of gaining highly aligned faces. This is demonstrate in Fig.~\ref{fig:meanfaces}, showing average faces from our frontalized LFW set (``LFW3D''), as well as funneled~\cite{huang2007unsupervised}, LFW-a~\cite{WHT:ACCV09:SSBS} (aligned using a commercial system), and deep-funneled~\cite{huang2012learning} versions of LFW. Our results all have slightly elongated faces, reflecting the shape of the reference face (Fig.~\ref{fig:process} (c)), yet are all clearly identifiable. Moreover, even when averaging the 530 LFW3D images of George W. Bush, our result retains crisp details and sharp edges despite the extreme variability of the original images, testifying to their aggressive alignment.

\section{Experiments}\label{sec:results}
Our method was implemented entirely in MATLAB, using the ``renderer'' function to render a reference view and produce the 2D-3D correspondences of Eq.~\ref{eq:proj_ref} and the ``calib'' function to estimate the projection matrix $\mathbf{C_Q}$, both functions available from~\cite{hassner2013viewing}. In all our experiments, we used the 3D face geometry used by~\cite{hassner2013viewing}, taken from the USF Human-ID database collection~\cite{usf_humanid}. Facial feature detection was performed using the SDM method~\cite{xiong2013supervised}, with their own implementation out-of-the-box. Its running time is approximately $.04$ seconds. Following detection, frontalization (including pose estimation) took an additional $\sim$0.1 seconds on $250\times 250$ pixel color images. These times measured on a standard Windows machine with an Intel i5 core processor and 8Gb RAM. Please see our project page for code and data:~\url{www.openu.ac.il/home/hassner/projects/frontalize}. 

\subsection{Qualitative results}\label{sec:qualitative}
Front-facing new views of Labeled Faces in the Wild images are provided throughout this paper. These were selected to show how our frontalization affects faces of varying age, gender, and ethnic backgrounds, as well as varying poses, occlusions, and more. We additionally compare our results with the two most relevant previous methods. Fig.~\ref{fig:occlusions} and~\ref{fig:surfacecompare} present results obtained using the code from~\cite{hassner2013viewing}. It was not designed specifically for frontalization, and so front facing views were manually produced. Fig.~\ref{fig:surfacecompare} additionally provides a comparison with~\cite{taigman2013deepface}.

\begin{table*}[!t]
\footnotesize{
\begin{center}
\begin{tabular}{lcccccc}
\toprule
		& \multicolumn{2}{c}{Funneled} & \multicolumn{2}{c}{LFW-a} & \multicolumn{2}{c}{LFW3D}\\
Method	& Values & Values Sqrt & Values & Values Sqrt	& Values & Values Sqrt \\ \hline
LBP & 0.6767  &  0.6782 & 0.6824 & 0.6790 &0.7465 $\pm$ 0.0053 (0.80) & 0.7322 $\pm$ 0.0061 (0.79) \\
TPLBP & 0.6875  &  0.6890 &0.6926& 0.6897 & 0.7502 $\pm$ 0.0055 (0.81) & 0.6723 $\pm$ 0.0323 (0.72) \\
FPLBP & 0.6865  &  0.6820 &0.6818& 0.6746 & 0.7265 $\pm$ 0.0143 (0.80) & 0.7345 $\pm$ 0.0061 (0.81) \\
OSS LBP & 0.7343   &  0.7463  & 0.7663 & 0.7820 & 0.8088 $\pm$ 0.0123 (0.87) & 0.8052 $\pm$ 0.0106 (0.87) \\
OSS TPLBP & 0.7163  &  0.7226 &0.7453  & 0.7514 & 0.8022 $\pm$ 0.0054 (0.87) & 0.7983 $\pm$ 0.0066 (0.87) \\
OSS FPLBP & 0.7175  &  0.7145 & 0.7466 & 0.7430 & 0.7852 $\pm$ 0.0057 (0.86) & 0.7822 $\pm$ 0.0049 (0.85) \\
Hybrid$^*$ & \multicolumn{2}{c}{0.7847 $\pm$ 0.0051}& \multicolumn{2}{c}{ 0.8255 $\pm$ 0.0031} & \multicolumn{2}{c}{0.8563 $\pm$ 0.0053~~(0.92)}  \\ \hline
Sub-SML~\cite{cao2013similarity} & \multicolumn{4}{c}{ 0.8973 $\pm$ 0.0038} & & \\
Sub-SML + Hybrid &\multicolumn{6}{c}{  0.9165 $\pm$ 0.0104 (0.92)} \\
\bottomrule
\end{tabular}
\vspace{-5mm}
\end{center}
\caption{{\bf Hybrid method verification results on the LFW benchmark.} Accuracy $\pm$ standard errors (SE) as well as area under the ROC curve (AUC) reported on the LFW View-2, restricted benchmark. Results for funneled and LFW-a images were taken from~\cite{WHT:ACCV09:SSBS}. No SE were reported for these methods. ``Value'' denotes the use of descriptor values directly (i.e., L2 or OSS distances); ``Values Sqrt'' represents Hellinger and Sqrt-OSS. $^*$ Results reported for funneled and LFW-a images were obtained using four representations and 16 similarity scores to our three and 12.}\label{tab:resultslfw}
}
\vspace{-3mm}
\end{table*}

\begin{figure}[t!]
\centering{
\includegraphics[width=.65\linewidth,clip,trim = 0mm 0mm 0mm 0mm]{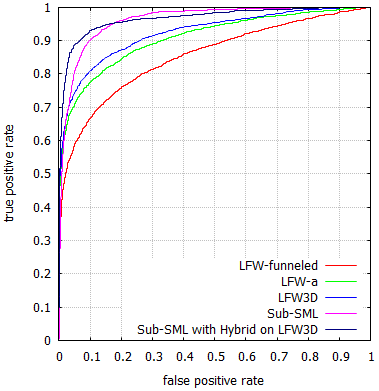}
}
\caption{{\bf ROC curves for LFW verification results.} Comparing the performance of the Hybrid method~\cite{WHT:ACCV09:SSBS} on Funneled LFW images, LFW-a and our own LFW3D, as well as the performance reported in Sub-SML~\cite{cao2013similarity}, and the accuracy obtained by combining both Sub-SML and Hybrid on LFW3D images.\vspace{-3mm}} \label{fig:LFWroc}
\end{figure}

\subsection{Face verification on the LFW benchmark}\label{sec:lfwtests}
We perform face verification tests on the Labeled Faces in the Wild (LFW) benchmark~\cite{lfw}. Its View-2 test protocol provides ten sets of 600 image pairs. Each one with 300 same person image pairs and 300 not-same pairs. Ten-fold cross validation tests are used taking each set in turn for testing and the rest for training, along with their ground truth same/not-same labels. Mean $\pm$ standard error (SE) over these ten folds are reported as well as area under the ROC curve (AUC). Our tests follow the ``Image-Restricted, Label-Free Outside Data'' protocol~\cite{lfwupdate}; outside data only used to train the facial feature detector.

We aim to see how much is face recognition performance improved with our frontalized faces. Thus, rather than using recent state-of-the-art methods which may mask the contribution of frontalization, we use the ``Hybrid'' method~\cite{WHT:ECCVW08:DBMW}, one of the first methods developed and successfully tested on LFW. Since then, newer, more modern methods have out-performed it with increasing margins by using better representations and learning techniques. We test how much of this performance gain can be reproduced by simply using better aligned images.

Our implementation uses these three representations: LBP~\cite{LBP2,ojala2002multiresolution}, TPLBP and FPLBP~\cite{WHT:ECCVW08:DBMW}. Image descriptors are compared using L2 distance, Hellinger distance~\cite{WHT:ECCVW08:DBMW} (L2 between descriptor element-wise square roots), One-Shot Similarity (OSS)~\cite{wolf2009one} and OSS applied to the descriptors' element-wise square root. In total, we use 3 descriptors $\times$ 4 similarities $=$ 12D vector of similarity values, classified by stacking~\cite{wolpert1992stacked} linear SVM classifiers.~\cite{svm}.

We frontalized LFW images as described in Sec.~\ref{sec:front}. Conditional symmetry (Sec.~\ref{sec:conditional}) was used here to also reject failed frontalizations: whenever six or more of the eight detectors failed on frontalized images, with and without soft-symmetry, the system defaulted to a planar alignment of the photo, in our case the corresponding deep-funneled images~\cite{huang2012learning}. Of the 13,233 LFW images, $\sim$2.5$\%$ were thus rejected, though more undetected failures exist. In most cases, these were due to occluded or extreme profile faces.

Results are compared to those reported on the LFW-a collection using a similar system~\cite{WHT:ACCV09:SSBS}. To our knowledge, these are the best results reported for the same face verification pipeline with an alternative alignment method, presumably optimized for best results. Alternatively, Deep-funneled images can be used instead of LFW-a but its performance gain over LFW-a are small~\cite{huang2012learning}.

Table~\ref{tab:resultslfw} lists our results and Fig.~\ref{fig:LFWroc} provides ROC curves. Evidently, our frontalized faces provide a performance boost of over 3$\%$ (possibly more, as the original Hybrid method included C1-Gabor descriptors in addition to the three we used). More importantly, these results show that rather than losing information when correcting pose using a single reference model, faces aligned this way are easier to classify than by using appearance preserving, in-plane alignment methods. 

We additionally report the performance obtained by combining the Sub-SML method of~\cite{cao2013similarity}, using their own implementation, with our Hybrid method, computed on frontalized LFW3D images. Sub-SML and Hybrid methods were combined by adding the Sub-SML image-pair similarity scores to the stacking SVM for a total of 13 values used for classification. For comparison, the performance originally reported by~\cite{cao2013similarity} is also provided. Adding the Hybrid method with LFW3D provides a 2$\%$ accuracy boost, raising the final performance to 0.9165 $\pm$ 1.04. To date, this is the highest score reported on the LFW challenge in the ``Image-Restricted, Label-Free Outside Data'' category.

\subsection{Gender estimation on the Adience benchmark}\label{sec:adiencetests}
The recently introduced Adience benchmark for gender estimation~\cite{eidinger2014age} has been shown to be the most challenging of its kind. It includes 26,580 photos of 2,284 subjects, downloaded from Flickr albums. Unlike LFW images, these images were automatically uploaded to Flickr from iPhone devices without manual filtering. They are thus far less constrained than LFW images. We use the {\em non}-frontal, version of this benchmark, which includes images of faces in $\pm$45$^\circ$ yaw poses. The test protocol defined for these images is 5-fold cross validation tests with album/subject-exclusive splits (images from the same subject or Flickr album appear in only one split). Performance is reported using mean classification accuracy $\pm$ standard errors (SE).

We compare results obtained by the best performing method in~\cite{eidinger2014age} on Adience images aligned with their proposed method with our implementation of the same method applied to frontalized Adience images (``Adience3D''). We again use LBP and FPLBP as image representations (results for TPLBP were not reported in~\cite{eidinger2014age}). Linear SVM were trained to classify descriptor vectors as belonging to either ``male'' or ``female'' using images in the training splits. We also tested training performed using ``dropout-SVM''~\cite{eidinger2014age} with a dropout rate of 0.5.

\begin{figure}[t!]
\centering{
\includegraphics[width=\linewidth,clip,trim = 0mm 3mm 0mm 0mm]{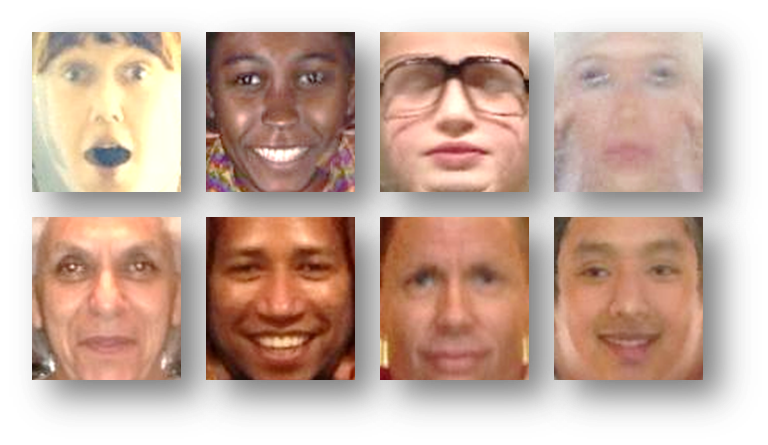}
}
\caption{{\bf Adience3D gender mis-classifications.} Top: Females classified as males; bottom: males classified as females. Errors result from the absence of clear gender related features or severely degraded images (e.g., top right).\vspace{-5mm}} \label{fig:adiencemistakes}
\end{figure}

Gender estimation results are listed in Table~\ref{tab:resultsadience}. Remarkably, frontalization advanced state-of-the-art performance by $\sim 4\%$. Some classification errors are additionally provided in Fig.~\ref{fig:adiencemistakes}. These demonstrate the elevated challenge of the Adience images along with successful frontalizations even with these challenging images.

\begin{table}[!t]
\footnotesize{
\begin{center}
\begin{tabular}{lcc}
\toprule
Method & Addience-aligned & Adience3D \\
\hline
LBP 									& 0.734 $\pm$ 0.007 & 0.800 $\pm$ 0.007\\
FPLBP 								& 0.726 $\pm$ 0.009 & 0.753 $\pm$ 0.010\\
LBP+FPLBP+Dropout 0.5 & 0.761 $\pm$ 0.009 & 0.793 $\pm$ 0.008\\ 
\bottomrule
\end{tabular}
\end{center}
\caption{{\bf Gender estimation on the Adience benchmark.} Mean accuracy ($\pm$ standard errors) reported on aligned Adience images~\cite{eidinger2014age} and our frontalized Adience3D images.\vspace{-3mm}}\label{tab:resultsadience}
}
\end{table}

\section{Conclusions}\label{sec:conc}
Computer vision systems have long since sought effective means of overcoming the many challenges of face recognition in unconstrained conditions. One of the key aspects of this problem is the variability of facial poses. Recently, an attractive, intuitive solution to this has been to artificially change the poses of faces appearing in photos, by generating novel, frontal facing views. This better aligns their features and reduces the variability that face recognition systems must address. 

We propose producing such frontalized faces using a simple yet, as far as we know, previously untested approach of employing a single 3D shape, unchanged, with all query photos. We show that despite the use of a face shape which can be very different from the true shapes, the resulting frontalizations lose little of their identifiable features. Furthermore, they are highly aligned, allowing for appearances to be easily compared across faces, despite possibly extreme pose differences in the input images. 

Beyond providing a simple and effective means for face frontalization, our work relates to a longstanding debate in computer vision on the role of appearances vs. 3D shape in face recognition. Our results seem to suggest that 3D information, when it is estimated directly from the query photo rather than provided by other means (e.g., stereo or active sensing systems), may potentially damage recognition performance instead of improving it. In the settings explored here, it may therefore be facial texture, rather than shape, that is key to effective face recognition.

{\small
\bibliographystyle{ieee}
\bibliography{frontalize}
}

\end{document}